\documentclass[letterpaper, 10 pt, conference]{ieeeconf}

\RequirePackage[normalem]{ulem} %DIF PREAMBLE
\RequirePackage{color}\definecolor{RED}{rgb}{1,0,0}\definecolor{BLUE}{rgb}{0,0,1} %DIF PREAMBLE
\usepackage[pdftex]{graphicx}
\graphicspath{{images}}
\DeclareGraphicsExtensions{.pdf,.jpeg,.png}

\usepackage{amsmath}
\usepackage{array}

\usepackage[font=small]{caption}
\usepackage{subcaption}
\usepackage{lipsum}
\usepackage{setspace}
\let\labelindent\relax
\usepackage{enumitem}
\usepackage{url}
\usepackage{amssymb}
\usepackage{bm}
\usepackage{graphicx}
\usepackage{float}
\usepackage[normalem]{ulem}
\usepackage{listings}
\usepackage{gensymb}
\usepackage{array}
\usepackage{pbox}
\usepackage{mathtools}
\usepackage{breqn}
\usepackage[ruled,vlined]{algorithm2e}
\usepackage{resizegather}
\usepackage{changepage}
\usepackage{titlesec}
\usepackage{nicefrac}
\usepackage{cases}
\usepackage{makecell}
\usepackage{nicematrix}
\usepackage{boldline}
\usepackage{xcolor}
\usepackage{xpatch}

\newcommand \reviewcomment[1]{\textcolor{red}{{#1}}}  % edits: keep these for final version
\usepackage[T3,T1]{fontenc}

\usepackage{hyperref}
\hypersetup{
     colorlinks   = true,
     linkcolor    = blue,
     citecolor    = red
}
\DeclareFontEncoding{LS1}{}{}
\DeclareFontSubstitution{LS1}{stix2}{m}{n}
\DeclareSymbolFont{letters-candra}{LS1}{stix2}{m}{it}
\DeclareMathAccent{\candra}{\mathalpha}{letters-candra}{"8B}

\titlespacing\section{0pt}{8pt plus 2pt minus 2pt}{5pt plus 2pt minus 2pt}
\titlespacing\subsection{0pt}{8pt minus 2pt}{5pt minus 2pt}
\newlength{\offsetpage}
{\end{adjustwidth}}
\xpretocmd{\algorithm}{\hsize=\linewidth}{}{}
\graphicspath{
    {images/}
}
\newcommand{\Exp}{\mathrm{Exp}}
\newcommand{\figref}{Fig.~\ref}
\newcommand{\tabref}{Table~\ref}

\newcommand{\mbs}{\bm}
\newcommand{\mbf}{\mathbf}
\newcommand{\bs}{\boldsymbol}

\renewcommand{\vec}{\mbox{vec}}
\newcommand{\bbm}{\begin{bmatrix}}
\newcommand{\ebm}{\end{bmatrix}}

\DeclareMathOperator*{\argmin}{\arg\!\min}

\DeclareSymbolFont{tipa}{T3}{cmr}{m}{n}
\DeclareMathAccent{\invbreve}{\mathalpha}{tipa}{16}

\newcolumntype{L}{>{$}l<{$}}

\newcounter{tableeqn}[table]

\newcounter{tablesubeqn}[tableeqn]

\newcommand \reviewtemp[1]{}
\newcommand \reviewdelmath[1]{}  % strikethroughs: won't need for final version
\newcommand \reviewdel[1]{}  % strikethroughs: won't need for final version

\newcommand\highlightReference[1]{%
  \expandafter\newcommand\csname highlightReference-#1\endcsname{}%
}
\let\oldbibitem\bibitem
\def\bibitem#1 #2\par{%
  \expandafter\ifx\csname highlightReference-#1\endcsname\relax
    \oldbibitem{#1}#2\par
  \else
    \oldbibitem{#1}\reviewcomment{#2}\par
  \fi
}

\makeatletter
\newcommand{\removelatexerror}{\let\@latex@error\@gobble}
\makeatother

\IEEEoverridecommandlockouts
\begin{document}

\title{\LARGE \bf KILO-EKF: Koopman-Inspired Learned Observations Extended Kalman Filter}

% \author{Author Names$^{1}$ redacted for anonymous review.
% \thanks{$^{1}$Author affiliations redacted for anonymous review.}
% }
\author{Zi Cong Guo$^{1}$, James R. Forbes$^{2}$, and Timothy D. Barfoot$^{1}$
\thanks{$^{1}$Zi Cong Guo and Timothy D. Barfoot are with the University of Toronto Robotics Institute, Toronto, Ontario, Canada (email: zc.guo@mail.utoronto.ca; tim.barfoot@utoronto.ca).}
\thanks{$^{2}$James R. Forbes is with the Department of Mechanical Engineering, McGill University, Montreal, Quebec, Canada (email: james.richard.forbes@mcgill.ca).}
}
% make the title area
\maketitle
\thispagestyle{empty}
\pagestyle{empty}

\begin{abstract}
We present the Koopman-Inspired Learned Observations Extended Kalman Filter (KILO-EKF), which combines a standard EKF prediction step with a correction step based on a Koopman-inspired measurement model learned from data. By lifting measurements into a feature space where they are linear in the state, KILO-EKF enables flexible modeling of complex or poorly calibrated sensors while retaining the structure and efficiency of recursive filtering. The resulting linear-Gaussian measurement model is learned in closed form from groundtruth training data, without iterative optimization or reliance on an explicit parametric sensor model. At inference, KILO-EKF performs a standard EKF update using Jacobians obtained via the learned lifting. We validate the approach on a real-world quadrotor localization task using an IMU, ultra-wideband (UWB) sensors, and a downward-facing laser. We compare against multiple EKF baselines with varying levels of sensor calibration. KILO-EKF achieves better accuracy and consistency compared to data-calibrated baselines, and significantly outperforms EKFs that rely on imperfect geometric models, while maintaining real-time inference and fast training. These results demonstrate the effectiveness of Koopman-inspired measurement learning as a scalable alternative to traditional model-based calibration.

% \vspace{10pt}
% {\textbf{\textit{Index Terms}---Localization; Probabilistic Inference.}}
\end{abstract}

\IEEEpeerreviewmaketitle
\section{Introduction}
State estimation is fundamental to many robotic systems, and recursive filtering methods are often required for real-time operation. Classical approaches, including the extended Kalman filter (EKF) and its variants~\cite{Simon2006}, rely on accurate system models to produce reliable state estimates. In practice, deriving precise analytical models is labor-intensive, system-specific, and sensitive to calibration~\cite{barfoot-txtbk}. When models are inaccurate or unavailable, model-based filters suffer from degraded accuracy or even divergence~\cite{probabilistic-robotics}. For many sensing modalities, such as ultra-wideband (UWB)~\cite{uwb-loc} and radio-frequency identification (RFID)~\cite{rfid-loc}, collecting data is relatively easy, while constructing faithful measurement models and associated noise characteristics remains a bottleneck. Environmental factors such as non-line-of-sight conditions~\cite{alarifi2016} may further complicate analytical modeling and calibration.

These challenges motivate a data-driven approach to modeling such complexities within an estimation framework. Prior work has applied Koopman operator theory~\cite{Koopman} to obtain data-driven representations of nonlinear systems. The framework has been used for system identification~\cite{koopman-dmdc,koopman-system-id}, control~\cite{koopman-control,control-affine-to-bilin,hagane2023}, and more recently for state estimation~\cite{kooplin,rckl}. However, the existing Koopman-based estimators have been formulated for vector-space states and operate in batch settings with constrained optimization, limiting their applicability to real-time filtering on Lie groups.

\begin{figure}[t]
    \centering
\includegraphics[width=0.99\columnwidth]{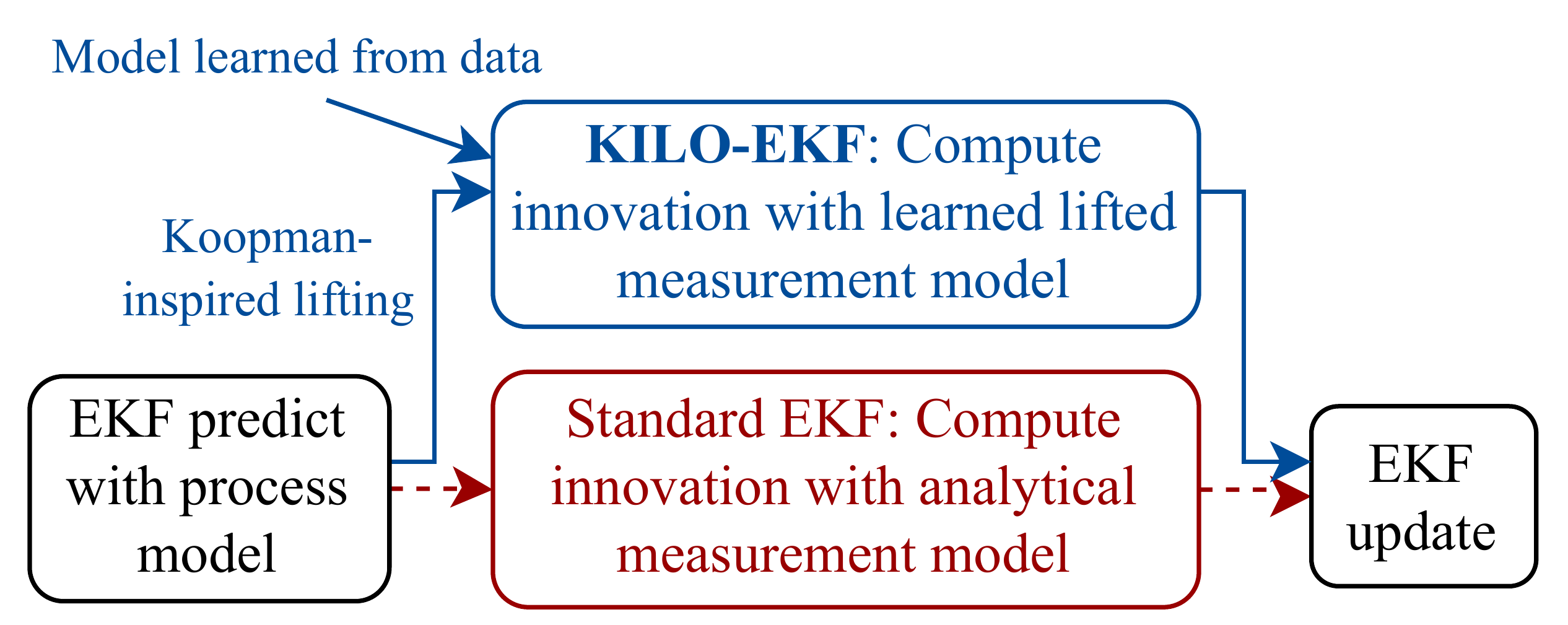}
\caption{Overview of the proposed KILO-EKF: the standard EKF structure is retained, while the measurement model is replaced with a Koopman-inspired, data-driven representation.}
\label{fig:kilo_ekf_diagram}
\end{figure}
\begin{figure}[t]
    \centering
\includegraphics[width=0.95\columnwidth, trim=2cm 4cm 0 2.5cm, clip]{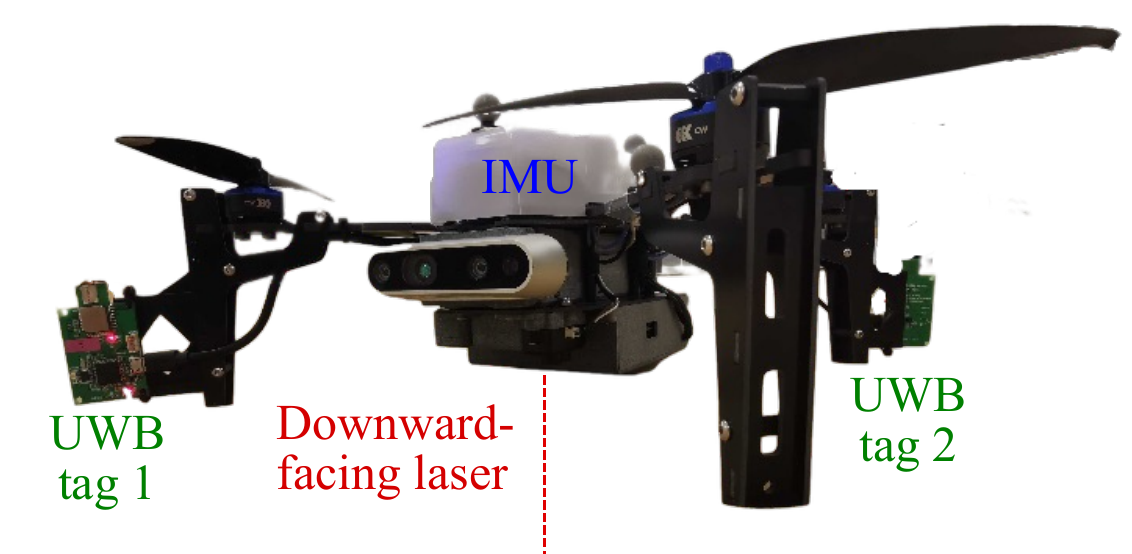}
\caption{System overview of the quadrotor platform~\cite{miluv} used to evaluate KILO-EKF, which learns data-driven measurement models for UWB ranging and laser height sensing.}
\label{fig:uwb_robot}
\end{figure}

We introduce the \emph{Koopman-Inspired Learned Observations EKF (KILO-EKF)}. The proposed method
\begin{itemize}
 \item learns a high-dimensional linear-Gaussian measurement model directly from data, without requiring explicit analytical measurement functions or Jacobians,
 \item integrates naturally into a standard EKF on Lie-group state spaces by replacing only the measurement model while preserving the recursive filtering structure,
 \item supports structured lifting through a combination of handcrafted features and systematic random feature expansions, and
 \item has a training cost that scales linearly with the amount of data, and an inference cost that is real-time and independent of training data size.
\end{itemize}

This paper is organized as follows. Section~\ref{sec:related-work} reviews related work on Koopman-based modeling and data-driven state estimation. Section~\ref{sec:preliminaries} introduces Koopman lifting for estimation and summarizes the standard EKF formulation on $\mathrm{SE}_2(3)$. Section~\ref{sec:koopman-ekf} presents the general KILO-EKF framework, and Section~\ref{sec:se23-koopman-ekf} specializes the framework to $\mathrm{SE}_2(3)$. Section~\ref{sec:experiments} experimentally validates KILO-EKF on a real-world dataset, and Section~\ref{sec:conclusion} concludes the paper.
% \vspace{-12pt}
\section{Related Work}
\label{sec:related-work}
Recent progress in data-driven modeling has enabled state estimation without explicit reliance on hand-crafted sensor models. One line of work employs kernel-based methods to represent nonlinear sensor behavior nonparametrically. Kernel mean embeddings~\cite{song} map probability distributions into reproducing kernel Hilbert spaces (RKHS), enabling estimators such as the kernel Bayes filter~\cite{kbr} and smoother~\cite{ksmoother}. Similarly, Gaussian process (GP) regression~\cite{gp-book} has been widely used to learn system and measurement models for state estimation~\cite{GPBayes}. While expressive, kernel methods typically incur cubic training complexity and require storing or approximating the training set at inference time, limiting real-time use without significant approximations~\cite{kkr}.

In contrast, Koopman-based methods~\cite{Koopman} represent nonlinear systems as linear dynamics in a lifted feature space, often enabling closed-form learning and inference with computational cost independent of training data size~\cite{dmd-big-book}. However, existing Koopman-based estimators either enforce nonlinear constraints~\cite{koopman-const-convex,koopman-nonlinear-se,rckl} or relax these constraints~\cite{kooplin,koopman-estimation-power-systems}, which can lead to inconsistent estimates~\cite{rckl}. Moreover, these methods are formulated for vector-space states. Although Koopman modeling has been explored on $\mathrm{SO}(3)$~\cite{koopman-so3} and $\mathrm{SE}(3)$~\cite{zinage2022,narayanan2023} for control, its application to state estimation on Lie groups remains limited.

Motivated by these gaps, we adopt a Koopman-inspired perspective for data-driven \emph{measurement} modeling within an EKF framework. The proposed method applies to general Lie-group state spaces (including vector spaces), avoids constrained optimization in the lifted space, and preserves the real-time performance of classical filtering.

\section{Preliminaries}
\label{sec:preliminaries}
\subsection{Koopman Lifting}
In this section, we briefly review Koopman lifting~\cite{Koopman} for discrete-time process models and discuss how a similar idea can be applied to measurement models. For comprehensive treatments of the Koopman operator, we refer the reader to~\cite{Koopman, dmd-big-book, mauroy_2020_koopman}.

Koopman lifting has traditionally been applied to process models, driven largely by its historical use in dynamical systems~\cite{koopman-control,koopman-dmdc}. The goal is to represent nonlinear dynamics as a linear evolution in a higher-dimensional space. Consider a noiseless discrete-time autonomous (i.e., no inputs) system, $\mbs{\xi}_k = \mbf{f}(\mbs{\xi}_{k-1})$, where $\mbs{\xi}_k$ denotes the state at timestep $k$ and $\mbf{f}(\cdot)$ is the process model. Then, there exists a lifting function, $\mbf{p}_{\mbs{\xi}}(\cdot)$, and a generally infinite-dimensional linear operator (the Koopman operator\footnote{In Koopman literature, the operator is canonically defined on scalar observables. Vector-valued liftings arise naturally when representing its action in a basis of observables.}), $\mbs{\mathcal{K}}$, such that $\mbf{p}_{\mbs{\xi}}(\mbs{\xi}_k) = (\mbs{\mathcal{K}} \mbf{p}_{\mbs{\xi}})(\mbs{\xi}_{k-1})$. Defining the lifted state $\mbf{x}_k = \mbf{p}_{\mbs{\xi}}(\mbs{\xi}_k)$ yields linear lifted dynamics of the form $\mbf{x}_k = \mbs{\mathcal{K}}\,\mbf{x}_{k-1}$.

In practice, several challenges arise. Exact finite-dimensional Koopman representations are rare, particularly for noisy systems, though truncated approximations can still be effective~\cite{koop-inv-with-sindy}. Moreover, real-world robotics systems include measurements, which must also be incorporated into the lifted representation for estimation.

Most existing work on Koopman-based estimation focuses on lifting the process model, or both the process and measurement models. For example, recent approaches~\cite{kooplin, rckl} lift both models for batch estimation in Euclidean state spaces. In contrast, we lift only the measurement model and embed it within a recursive EKF framework applicable to Lie groups such as $\mathrm{SE}_2(3)$. This perspective is well suited to robotics, where process models are often known, while measurement models and their associated noise characteristics can be more difficult to derive, particularly on Lie groups.

\subsection{Standard Discrete-Time EKF on SE\textsubscript{2}(3)}
\label{subsec:se23-ekf}
In this section, we briefly summarize the standard formulation for discrete-time EKF on $\mathrm{SE}_2(3)$ and refer the reader to~\cite{miluv} for more details. A general robotics system consists of a process model and a measurement model:
\begin{subequations}\label{eq:general-model}
\begin{align}
 \mbs{\xi}_k &= \mbf{f}(\mbs{\xi}_{k-1}, \mbs{\nu}_k, \mbs{\epsilon}_k), \label{eq:general-model-1} \\
 \mbs{\gamma}_k &= \mbf{g}(\mbs{\xi}_k, \mbs{\eta}_k), \label{eq:general-model-2}
\end{align}
\end{subequations}
where $\mbs{\xi}_k$ is the system state at time $k$, $\mbs{\nu}_k$ is the control or input measurement, $\mbs{\epsilon}_k$ is the process noise, $\mbs{\gamma}_k$ is the measurement, and $\mbs{\eta}_k$ is the measurement noise.

Many robotic systems operating in three-dimensional space, such as aerial vehicles and legged platforms, are equipped with an inertial measurement unit (IMU). From the bias-corrected IMU readings, we construct the input $\mbs{\nu}_k$, consisting of angular velocity $\mbs{\omega}_k$ and specific force $\mbf{a}_k$. These interoceptive measurements, together with additional exteroceptive sensors, are commonly fused using an EKF on the Lie group $\mathrm{SE}_2(3)$. Accordingly, we define the EKF state as $\mbs{\xi}_k = \big\{\mbf{T}_k,\; \mbf{b}_k\big\} \in \mathrm{SE}_2(3) \times \mathbb{R}^6$, where
\begin{gather}
\label{eq:se23-state}
\mbf{T}_k = \bbm
\mbf{C}_k & \mbf{v}_k & \mbf{t}_k \\
\mbf{0} & 1 & 0 \\
\mbf{0} & 0 & 1
\ebm \in \mathrm{SE}_2(3), \quad \mbf{b}_k = \bbm
{\mbf{b}^\omega_k} \\ {\mbf{b}^a_k} \ebm \in \mathbb{R}^6,
\end{gather}
where $\mbf{C}_k \in \mathrm{SO}(3)$ is the body-to-world rotation, $\mbf{v}_k$ and $\mbf{t}_k$ are the velocity and position in the world frame, and $\mbf{b}_k$ is the IMU biases. We define the error state using a right perturbation on $\mathrm{SE}_2(3)$. 
Specifically, the true state $\mbs{\xi}_k$ is related to a nominal state $\bar{\mbs{\xi}}_k$ by $\mbf{T}_k = \bar{\mbf{T}}_k \Exp(\delta\mbs{\zeta}_k)$, $
\mbf{b}_k = \bar{\mbf{b}}_k + \delta\mbf{b}_k$, 
where the $15$-dimensional error state is
\begin{gather}
\label{eq:se23-error-state}
\delta \mbs{\xi}_k =
\bbm
\delta\mbs{\zeta}_k^\top &
\delta\mbf{b}_k^\top
\ebm^\top, \quad
\delta\mbs{\zeta}_k =
\bbm
\delta\mbs{\theta}_k^\top &
\delta\mbf{v}_k^\top &
\delta\mbf{t}_k^\top
\ebm^\top.
\end{gather}
Here, $\Exp(\cdot) = \exp((\cdot)^\wedge)$, where $(\cdot)^\wedge$ denotes the mapping from $\mathbb{R}^9$ to the Lie algebra $\mathfrak{se}_2(3)$~\cite{micro-lie-theory}. The nominal state is propagated using a deterministic process model: $\bar{\mbf{T}}_k =
\bar{\mbf{T}}_{k-1}
\Exp\!\left(
\mbs{\nu}_k \Delta t_k
\right)$, $\bar{\mbf{b}}_k =
\bar{\mbf{b}}_{k-1}$.

The EKF assumes zero-mean Gaussian process and measurement noises: $\mbs{\epsilon}_k \sim \mathcal{N}(\mbf{0}, \mbf{Q}_k)$ and $\mbs{\eta}_k \sim \mathcal{N}(\mbf{0}, \mbf{R}_{\gamma,k})$. Process noise is modeled through the linearized error-state dynamics, $\delta \mbs{\xi}_k
=
\mbf{F}_{k-1}\,\delta \mbs{\xi}_{k-1}
+
\mbf{G}_k\,\mbs{\epsilon}_k$. IMU noise enters through $\mbs{\epsilon}_k$, and the biases are modeled as a random walk.

The EKF maintains a Gaussian belief over the error state with covariance $\mbf{P}_k$. 
During the prediction step, the nominal state and covariance are propagated, where the covariance prediction is given by $\check{\mbf{P}}_{k} =
\mbf{F}_{k-1}\,\hat{\mbf{P}}_{k-1}\,\mbf{F}_{k-1}^\top
+
\mbf{G}_k\,\mbf{Q}_k\,\mbf{G}_k^\top$, yielding the prior
$(\check{\mbs{\xi}}_k, \check{\mbf{P}}_k)$. Here, $\check{(\cdot)}$ and $\hat{(\cdot)}$ denote prior and posterior quantities, respectively.

For the measurement update, we receive an exteroceptive measurement $\mbs{\gamma}_{k}$, related to the state by the nonlinear measurement model in~\eqref{eq:general-model-2}. The model, $\mbf{g}(\cdot)$, is typically derived from sensor geometry, calibration parameters, and prior knowledge of the environment. Using the predicted state, $\check{\mbs{\xi}}_{k}$, the EKF computes the innovation $\mbs{r}_{k} = \mbs{\gamma}_{k} - \mbf{g}(\check{\mbs{\xi}}_{k},\mbf{0})$, and linearizes the measurement model with respect to the error state to obtain the Jacobian, $\mbf{H}_{k} = \left.
\frac{\partial \delta \mbs{\gamma}}{\partial \delta \mbs{\xi}}
\right|_{\delta\mbs{\xi}=\mbf{0}}$. 
% See Appendix~\ref{app:jacobian} for details on the Jacobian derivation for $\mathrm{SE}_2(3)$. % For arXiv submission
Then, the Kalman gain is $\mbf{K}_{k} = \check{\mbf{P}}_{k}\mbf{H}_{k}^\top
\big(\mbf{H}_{k}\check{\mbf{P}}_{k}\mbf{H}_{k}^\top
+
\mbf{R}_{\gamma,k}
\big)^{-1}$, and thus the correction is $\delta\hat{\mbs{\xi}}_{k} = \mbf{K}_{k}\mbs{r}_{k}$. Writing $\delta\hat{\mbs{\xi}}_{k} = \bbm \delta\hat{\mbs{\zeta}}_{k}^\top & \delta\hat{\mbf{b}}_{k}^\top \ebm^\top$, we update via $\hat{\mbf{T}}_{k} = \check{\mbf{T}}_k \Exp(\delta\hat{\mbs{\zeta}}_{k})$ and $\hat{\mbf{b}}_{k} = \check{\mbf{b}}_{k} + \delta\hat{\mbf{b}}_{k}$, followed by the standard EKF covariance update to obtain $\hat{\mbf{P}}_{k}$.

Having established the standard EKF as our baseline, we next introduce the KILO-EKF, which adopts a Koopman-inspired approach for learning measurement models while retaining the classical prediction step.

\section{The General KILO-EKF}
\label{sec:koopman-ekf}
In this section, we present KILO-EKF's general framework for mapping nonlinear measurement functions into a linear representation in a lifted feature space. The resulting formulation is agnostic to the underlying state space and filtering architecture. We will specialize this approach to $\mathrm{SE}_2(3)$ in the next section by defining appropriate lifting functions for the states.

By lifting measurement model~\eqref{eq:general-model-2}, system~\eqref{eq:general-model} can be represented as
\begin{subequations}
\label{eq:lifted-model}
\begin{align}
 \mbs{\xi}_k &= \mbf{f}(\mbs{\xi}_{k-1}, \mbs{\nu}_k, \mbs{\epsilon}_k), \label{eq:lifted-model-1} \\
 \mbf{y}_k &= \mbf{D} \mbf{x}_k + \mbf{n}_k, \label{eq:lifted-model-2}
\end{align}
\end{subequations}
where for some nonlinear lifting functions $\mbf{p}_{\mbs{\xi}}(\cdot)$ and $\mbf{p}_{\mbs{\gamma}}(\cdot)$, the lifted quantities are $\mbf{x}_k,\mbf{n}_k \in \mathcal{X}$ and $\mbf{y}_k \in \mathcal{Y}$, where
\begin{gather}
\label{eq:lifted-quantities}
\mbf{x}_k = \mbf{p}_{\mbs{\xi}}(\mbs{\xi}_k), \quad
\mbf{y}_k = \mbf{p}_{\mbs{\gamma}}(\mbs{\gamma}_k), \quad
 \mbf{n}_k \sim \mathcal{N}(\mbf{0}, \mbf{R}).
\end{gather}
We have modeled the lifted model as linear Gaussian. While this is approximate, the aggregation of multiple error sources in high-dimensional feature spaces can often be reasonably approximated as Gaussian under the Central Limit Theorem~\cite{vershynin_hdp}. Moreover, a linear-Gaussian formulation yields a tractable measurement update and is consistent with the assumptions underlying the EKF. Prior work on Koopman-based estimation has also demonstrated strong empirical performance under this assumption~\cite{kooplin, rckl}.

In general, lifting the measurements through $\mbf{y}_k = \mbf{p}_{\mbs{\gamma}}(\mbs{\gamma}_k)$ is optional, as sufficiently expressive state liftings can capture nonlinear measurement relationships. However, appropriate measurement liftings can simplify the learned model and improve numerical conditioning in practice. See Section~\ref{subsec:kilo-ekf-implementation} for an example. For now, we turn to learning the lifted model parameters from data for use in estimation.

\subsection{Learning the Lifted Measurement Model}
Our objective is to learn $\mbf{D}$ and $\mbf{R}$ in~\eqref{eq:lifted-model} and~\eqref{eq:lifted-quantities} from groundtruth data. Suppose we have a dataset consisting of $P$ states and measurements: $\{ \mbs{\xi}^{(i)}, \mbs{\gamma}^{(i)} \}_{i=1}^P$. We write the original data in block-matrix form as $\mbs{\Xi} = \bbm \mbs{\xi}^{(1)} & \cdots & \mbs{\xi}^{(P)} \ebm$ and $\mbs{\Gamma} = \bbm \mbs{\gamma}^{(1)} & \cdots & \mbs{\gamma}^{(P)} \ebm$.

The data translates to $\{ \mbf{x}^{(i)}, \mbf{y}^{(i)} \}_{i=1}^P$ in the lifted space such that $\mbf{y}^{(i)} = \mbf{D} \mbf{x}^{(i)} + \mbf{n}^{(i)}$ for some unknown noise, $\mbf{n}^{(i)} \sim \mathcal{N}(\mbf{0}, \mbf{R})$. Then, we can write the lifted matrix form for the measurement model as $\mbf{Y} = \mbf{D}\mbf{X} + \mbf{N}$, where $\mbf{X} = \begin{bmatrix} \mbf{x}^{(1)} & \cdots & \mbf{x}^{(P)} \end{bmatrix}$,
 $\mbf{Y} = \begin{bmatrix} \mbf{y}^{(1)} & \cdots &\mbf{y}^{(P)} \end{bmatrix}$, and $\mbf{N} = \begin{bmatrix} \mbf{n}^{(1)} & \cdots & \mbf{n}^{(P)} \end{bmatrix}$.

We now formulate a loss function for learning the system matrices from data. Rather than EDMD-based formulations involving spectral decompositions~\cite{dmd-book}, we adopt a Tikhonov-regularized approach that admits a closed-form solution. Specifically, we cast the learning problem as a maximum a posteriori (MAP) estimation with conjugate priors on 
$\mbf{D}$ and $\mbf{R}$. The resulting optimization problem is
\begin{gather}
 \left\{\mbf{D}^\star,\mbf{R}^\star\right\} = \argmin_{\{\mbf{D},\mbf{R}\}} V(\mbf{D},\mbf{R}),
\end{gather}
where the loss function is $V=V_1+V_2$ with 
\begin{subequations}
\begin{align}
% V_1
V_1 =& \frac{1}{2} \left\| \mbf{Y} - \mbf{D} \mbf{X} \right\|^2_{\mbf{R}^{-1}} + \frac{1}{2} {P} \ln \left| \mbf{R} \right|, \\
% V_2
V_2 =& \frac{1}{2} {P} \tau_D \left\| \mbf{D} \right\|^2_{\mbf{R}^{-1}} + \frac{1}{2} {P} \tau_R \, \mbox{tr}(\mbf{R}^{-1}).
\end{align}
\end{subequations}
The norm is a weighted Frobenius matrix norm: $\left\| \mbf{X} \right\|_{\mbf{W}} = \sqrt{\mbox{tr}\left( \mbf{X}^T \mbf{W} \mbf{X} \right)}$. $V_1$ corresponds to the negative log-likelihood of the data under Gaussian noise, ignoring the normalizing constant. $V_2$ corresponds to the negative log-priors on the model parameters: a Tikhonov prior on $\mbf{D}$ and an (isotropic) inverse-Wishart (IW) prior on $\mbf{R}$. IW distributions have been demonstrated to be robust priors for learning covariances \cite{esgvi-extended}. We tune the regularizing hyperparameters, $\tau_D$ and $\tau_R$, from data using cross-validation.

We find the critical points by setting $\frac{\partial V}{\partial \mbf{D}}$ and $\frac{\partial V}{\partial \mbf{R}^{-1}}$ to zero, yielding
\begin{subequations}\label{eq:abhcqr}
\begin{align}
\mbf{D} & = (\mbf{Y}\mbf{X}^T) (\mbf{X}\mbf{X}^T + {P} \tau_D \mbf{1})^{-1}, \\
\mbf{R} & = \frac{1}{{P}} (\mbf{Y} - \mbf{D} \mbf{X})(\mbf{Y} - \mbf{D} \mbf{X})^T + \tau_D \mbf{D} \mbf{D}^T +  \tau_R \mbf{1},
\end{align}
\end{subequations}
where $\mbf{1}$ represents the identity operator for the appropriate domains. This procedure is one shot and linear in the amount of training data, $P$, for both computation and storage.

\subsection{Inference}
After learning the lifted measurement model parameters $\mbf{D}$ and $\mbf{R}$, we use them for state estimation within an EKF framework. In this setting, inference proceeds by computing the measurement Jacobian with respect to the original state. Using the lifted measurement model, this Jacobian can be written using the chain rule as
\begin{gather}
\label{eq:koopman-jacobian}
\mbf{H}_k
=
\frac{\partial \delta \mbf{y}}{\partial \delta \mbs{\xi}}
\biggr\rvert_{\delta \mbs{\xi}=\mbf{0}}
=
\mbf{D}
\frac{\partial \delta \mbf{p}_{\mbs{\xi}}}{\partial \delta \mbs{\xi}}
\biggr\rvert_{\delta \mbs{\xi}=\mbf{0}}.
\end{gather}
This expression is straightforward to compute, as the lifting function $\mbf{p}_{\mbs{\xi}}(\cdot)$ is chosen a priori and is differentiable by construction. With this Jacobian, the EKF measurement update follows the standard formulation using the learned covariance $\mbf{R}$ and leaving the prediction step unchanged\footnote{The proposed measurement-learning component is independent of the specific filtering formulation and can be integrated within alternative variants, such as the invariant EKF.}. With multiple sensors, we learn a separate lifted measurement model for each sensor, then apply the corresponding measurement updates sequentially, as in standard multi-sensor EKFs.

A convenient and widely used choice for the lifting function is to include the original state itself. Specifically, the state, $\mbs{\xi}_k$, is concatenated with nonlinear features, $\tilde{\mbf{p}}_{\mbs{\xi}}(\mbs{\xi}_k)$, as
\begin{gather}
\mbf{p}_{\mbs{\xi}}(\mbs{\xi}_k)
=
\bbm
\mbs{\xi}_k^\top &
\tilde{\mbf{p}}_{\mbs{\xi}}(\mbs{\xi}_k)^\top
\ebm^\top .
\end{gather}
This structure introduces only a modest increase in dimension, while enabling direct recovery of the original state.

This inference procedure differs from prior Koopman-based batch estimators~\cite{kooplin, rckl}, where the optimization variable is the lifted state, $\mbf{x}_k$. In those approaches, consistency between the lifted variables and the original state must be enforced explicitly through nonlinear equality constraints~\cite{rckl} of the form $\mbf{h}(\mbf{x}_k) = \tilde{\mbf{x}}_k - \tilde{\mbf{p}}_{\mbs{\xi}}(\mbs{\xi}_k) = \mbf{0}$, where $\mbf{x}_k = \bbm \mbs{\xi}_k^\top & \tilde{\mbf{x}}_k^\top \ebm^\top$. These constraints ensure that $\hat{\mbf{x}}_k \in \mathcal{X}$. That is, the estimator's solution must lie on the manifold induced by the lifting map. In batch estimation, such constraints are a natural modeling choice, trading nonlinear model structure for explicit constraints with favorable numerical properties~\cite{nocedal, dellaert_kaess}. In a recursive filtering setting, however, this trade-off offers little benefit. In KILO-EKF, the original state $\mbs{\xi}_k$ is estimated directly, and the lifted state is generated deterministically as $\mbf{x}_k=\mbf{p}_{\mbs{\xi}}(\mbs{\xi}_k)$. Consequently, consistency is automatically satisfied, and the measurement Jacobian is obtained by applying the chain rule. This avoids constrained optimization at inference time while preserving the standard EKF structure required for real-time operation.

\section{KILO-EKF in $\mathrm{SE}_2(3)$}
\label{sec:se23-koopman-ekf}
Having established the general framework of KILO-EKF, we now specialize to $\mathrm{SE}_2(3)$. We first define suitable lifting functions for the state, followed by the computation of the measurement Jacobian required for the EKF update.

\subsection{Lifting Functions}
\label{subsec:se23-lifting}
In the following, we make explicit the construction of the lifting function
$\mbf{p}_{\mbs{\xi}}(\cdot)$. From Section~\ref{subsec:se23-ekf}, our state is $\mbs{\xi}_k = \{\mbf{T}_k, \mbf{b}_k\}$, as defined in~\eqref{eq:se23-state}. To apply Euclidean feature maps, we introduce $\mbf{s}_k$, a vector representation of our EKF state via a mapping $\mbf{q}(\cdot): \mathrm{SE}_2(3) \times \mathbb{R}^6 \to \mathbb{R}^{15}$, where
\begin{gather}
\label{eq:se23-state-vector}
 \mbf{s}_k = \mbf{q}(\mbs{\xi}_k) =
 \bbm \vec(\mbf{C}_k)^\top & \mbf{t}_k^\top & \mbf{v}_k^\top & \mbf{b}_{a,k}^\top & \mbf{b}_{\omega,k}^\top \ebm^\top.
\end{gather}
We then define the lifting function $\mbf{p}_{\mbs{\xi}}(\cdot)$ as a function of $\mbf{s}_k$.

For lifting functions, we use squared-exponential random Fourier features (SERFFs)~\cite{rff}, which map vector-space inputs to a finite-dimensional set of sinusoidal features. Given an input $\mbf{s}$, the SERFF mapping is defined as
\begin{subequations}
\label{eq:serff}
\begin{align}
\mbf{z}(\mbf{s}) &=
\bbm
 \frac{1}{\sqrt{R}} \mbf{z}_{\bs{\varpi}_1}(\mbf{s})^\top &
 \cdots &
 \frac{1}{\sqrt{R}} \mbf{z}_{\bs{\varpi}_R}(\mbf{s})^\top
\ebm^\top, \\
\mbf{z}_{\bs{\varpi}_i}(\mbf{s}) &=
\bbm
 \sqrt{2} \cos(\bs{\varpi}_i^\top \mbf{s}) &
 \sqrt{2} \sin(\bs{\varpi}_i^\top \mbf{s})
\ebm^\top,
\end{align}
\end{subequations}
where the frequency vectors $\bs{\varpi}_i$ are drawn independently as
$\bs{\varpi}_i \sim \mathcal{N}(\mbf{0}, \mbf{W}/k_\ell)$.
Here, $\mbf{W}$ is a (typically diagonal) weighting matrix that controls the relative importance of each input dimension, and $k_\ell$ is a length-scale parameter. Both hyperparameters can be selected a priori or tuned from data.

SERFFs are commonly used as lifting functions in Koopman-based methods due to their strong empirical performance~\cite{koopman-with-rff}, and have been successfully applied in prior Koopman-based estimation work~\cite{kooplin, rckl}. From a theoretical perspective, as the number of random features increases, SERFFs asymptotically approximate the squared-exponential kernel~\cite{rff-koopman, kooplin}. Since this kernel is universal, it can approximate a broad class of nonlinear measurement functions~\cite{steinwart2001}. Despite this expressiveness, kernel methods scale poorly, with computational complexity typically growing cubically in the number of training samples~\cite{scholkopf2002}. In contrast, SERFFs offer a more flexible alternative, as the trade-off between approximation accuracy and computational cost can be directly controlled by selecting the number of random features.

The proposed lifting framework can be applied directly without modification by defining $\mbf{p}_{\mbs{\xi}}(\mbf{s}) = \mbf{z}(\mbf{s})$ from~\eqref{eq:serff}. In many practical scenarios, however, some prior knowledge of the system is available. In such cases, it is beneficial to incorporate this knowledge by constructing handcrafted features inspired by known measurement models. Handcrafted features can explicitly encode known structure and reduce the number of random features required. See Section~\ref{sec:experiments} for an example. It is also often advantageous to restrict the lifting to a subset of the original state variables. For example, if the measurement is known to depend only on the pose, we modify~\eqref{eq:se23-state-vector} to retain only the corresponding components and discard the remaining state variables:
\begin{gather}
    \label{eq:reduced-state}
 \mbf{s}^\prime_k = \bbm \vec(\mbf{C}_k)^\top & \mbf{t}_k^\top \ebm^\top .
\end{gather}
This reduces the dimensionality of the input to the lifting function and, consequently, the number of lifted features required. The resulting lifting function is then defined as
\begin{gather}
\label{eq:reduced-state-lifted}
 \mbf{p}_{\mbs{\xi}}(\mbf{s}_k^\prime) =
 \bbm
 {\mbf{s}_k^\prime}^\top &
 \mbf{h}(\mbf{s}_k^\prime)^\top &
 \mbf{z}(\mbf{s}_k^\prime)^\top
 \ebm^\top,
\end{gather}
where $\mbf{h}(\cdot)$ is a handcrafted feature function. This design enables flexible incorporation of domain knowledge while preserving the generality of the KILO-EKF framework.
\begin{figure*}[t]
  \centering
  \vspace{5pt}
  % Titles row
    \begin{minipage}[t]{0.19\textwidth}\centering \strut Fold 1\end{minipage}\hfill
    \begin{minipage}[t]{0.19\textwidth}\centering \strut Fold 2\end{minipage}\hfill
    \begin{minipage}[t]{0.19\textwidth}\centering \strut Fold 3\end{minipage}\hfill
    \begin{minipage}[t]{0.19\textwidth}\centering \strut Fold 4\end{minipage}\hfill
    \begin{minipage}[t]{0.19\textwidth}\centering \strut Fold 5\end{minipage}
    \vspace{2pt}

  \begin{subfigure}{0.19\textwidth}
    \includegraphics[width=\linewidth]{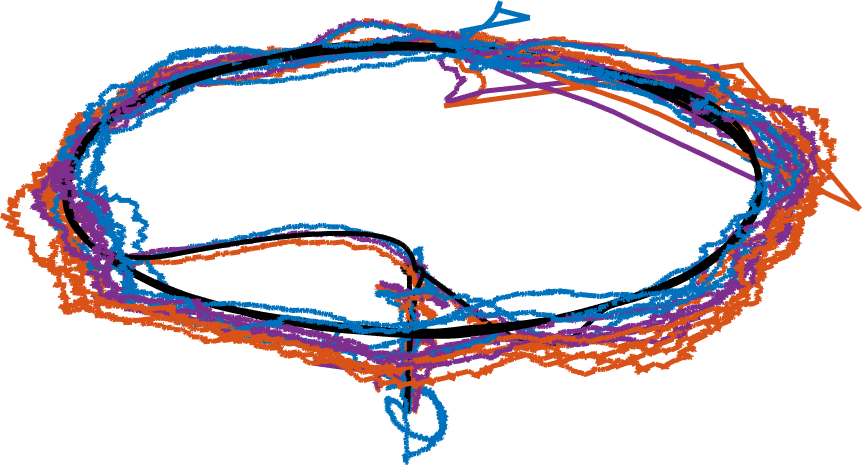}
  \end{subfigure}
  \begin{subfigure}{0.19\textwidth}
    \includegraphics[width=\linewidth]{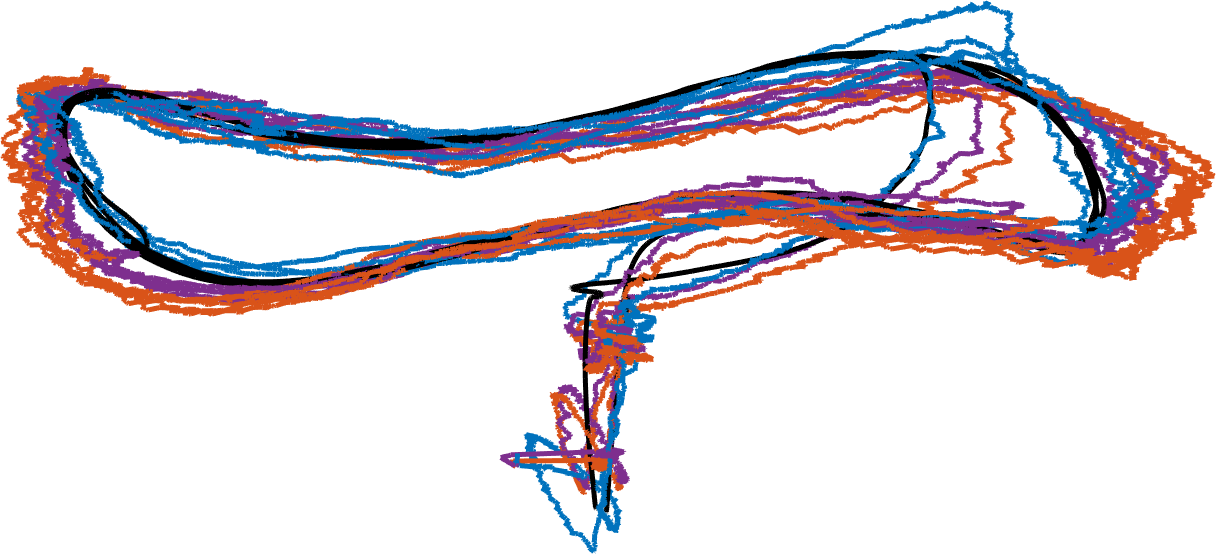}
  \end{subfigure}
  \begin{subfigure}{0.19\textwidth}
    \includegraphics[width=\linewidth]{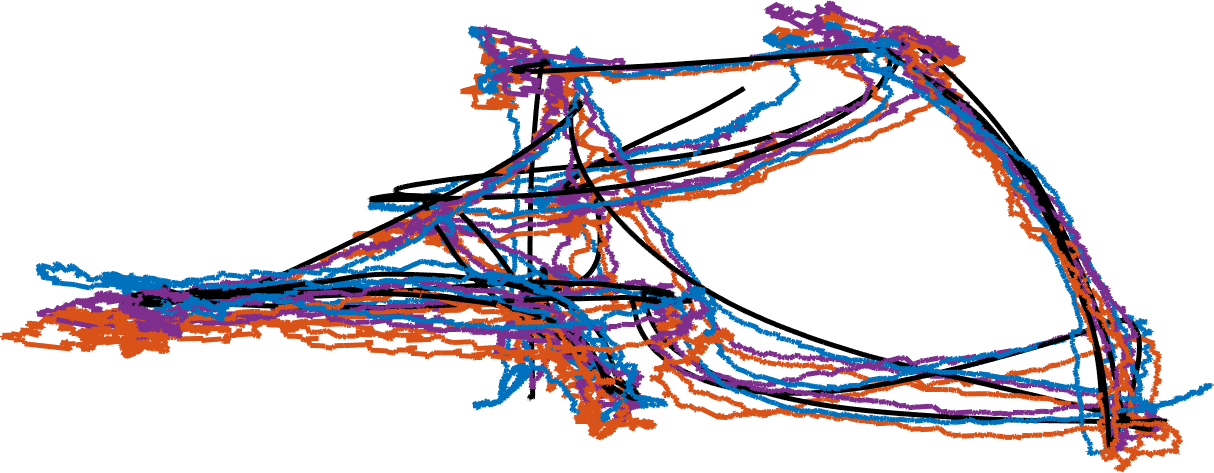}
  \end{subfigure}
  \begin{subfigure}{0.19\textwidth}
    \includegraphics[width=\linewidth]{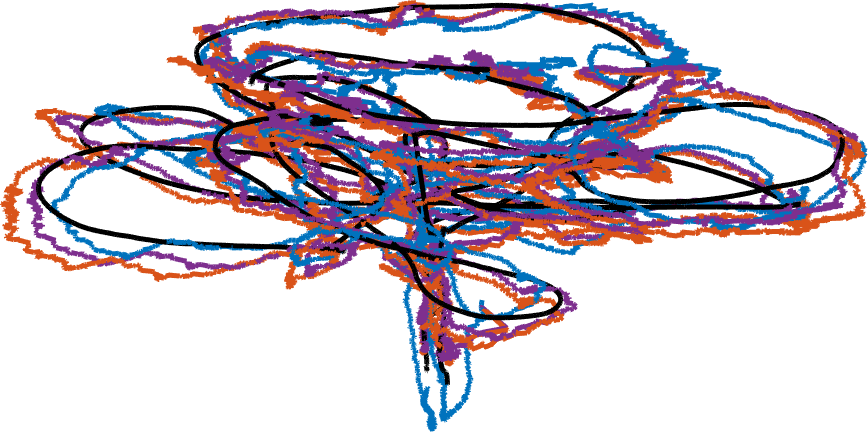}
  \end{subfigure}
  \begin{subfigure}{0.19\textwidth}
    \includegraphics[width=\linewidth]{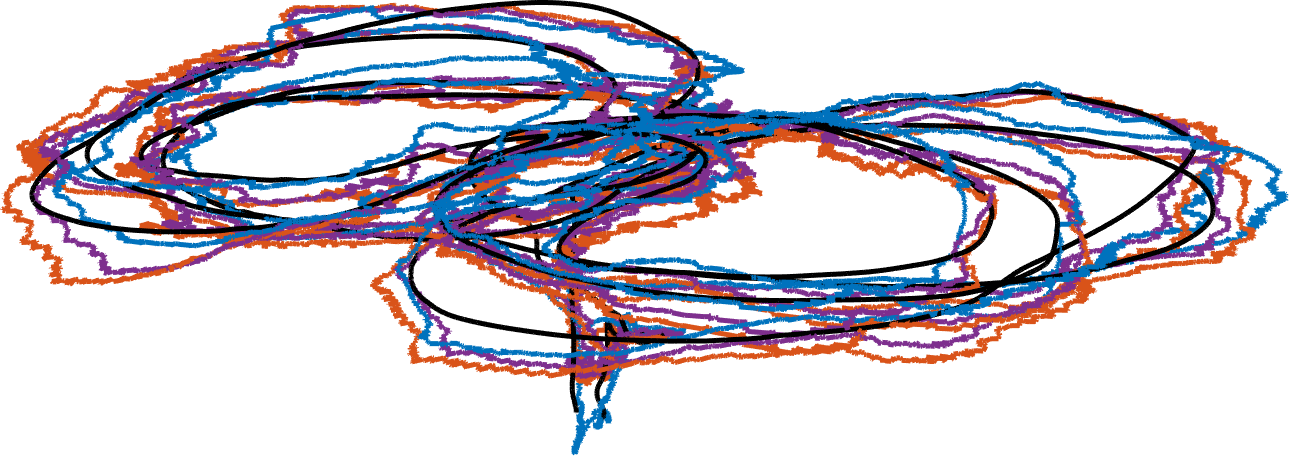}
  \end{subfigure}

  \vspace{1mm}
  \includegraphics[width=0.7\textwidth]{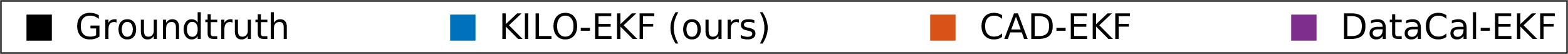}

  \caption{Trajectory estimates for the five cross-validation folds, illustrating the diversity of path geometries used in evaluation. Qualitatively, although not always visually pronounced, the proposed KILO-EKF generally tracks the groundtruth more closely than the two baselines.}
  \label{fig:koopekf_trajs}
\end{figure*}

\subsection{Measurement Jacobian}
Using the lifted measurement model, the EKF measurement Jacobian follows directly from the chain rule as in~\eqref{eq:koopman-jacobian}. The geometric aspects of the $\mathrm{SE}_2(3)$ linearization are identical to those of a conventional EKF.
% and are summarized in Appendix~\ref{app:jacobian}. % For arXiv submission
For the SERFF lifting functions defined in~\eqref{eq:serff}, the required derivatives are available in closed form:
\begin{gather}
\frac{\partial \mbf{z}_{\bs{\varpi}_i}}{\partial \mbf{s}}
=
\bbm
-\sqrt{2}\sin(\bs{\varpi}_i^\top \mbf{s}) \\
\phantom{-}\sqrt{2}\cos(\bs{\varpi}_i^\top \mbf{s})
\ebm
\bs{\varpi}_i^\top.
\end{gather}

\section{Experiments}
\label{sec:experiments}

\subsection{Problem Setup}
We validate the proposed KILO-EKF on a real-world 3D localization dataset named MILUV~\cite{miluv}. A quadrotor operates in a $3\,\mathrm{m} \times 3\,\mathrm{m} \times 2\,\mathrm{m}$ indoor environment, with an onboard IMU providing linear acceleration and angular velocity measurements. For exteroceptive sensing, the quadrotor carries two UWB tags and receives range measurements from six fixed UWB anchors. In addition, a downward-facing laser provides measurements of the vertical height.
%\footnote{Laser height measurements are necessary for these datasets, as the system is only weakly observable otherwise.}
See \figref{fig:uwb_robot} for a visualization. The localization objective is to estimate the quadrotor's pose trajectory using IMU, UWB, and laser measurements. Groundtruth poses are obtained from a motion-capture system and are used for supervised learning and evaluation under a cross-validation protocol.

\subsection{EKF Baselines}
\label{sec:baseline_ekf}
Our baseline estimators are standard $\mathrm{SE}_2(3)$ EKFs implemented as part of the MILUV dataset~\cite{miluv}. For timestamp $k$, UWB tag $i$, and anchor $j$, the analytical UWB range measurement model is
\begin{gather}
    \label{eq:uwb-model}
 {\gamma}_{i,j,k}
 =
 {g}_{i,j}(\mbs{\xi}_k)
 =
 \left\lVert
 \mbf{r}_{\ell,j}
 -
 \mbf{t}_k
 -
 \mbf{C}_k \mbf{r}_{b,i}
 \right\rVert
 +
 {\eta}_{i,j,k},
\end{gather}
where $\mbf{r}_{\ell,j}$ denotes the position of anchor $j$ expressed in the world frame, $\mbf{r}_{b,i}$ denotes the position of UWB tag $i$ expressed in the vehicle frame, and ${\eta}_{i,j,k}$ is zero-mean measurement noise.

In addition to UWB range measurements, we have a downward-facing laser that measures the vertical height. The analytical laser measurement model is
\begin{gather}
 \gamma_{L,k}
 =
 g_L(\mbs{\xi}_k)
 =
 t_{z,k}
 +
 \eta_{L,k},
\end{gather}
where $t_{z,k}$ denotes the $z$-component of the position $\mbf{t}_k$, and $\eta_{L,k}$ is zero-mean measurement noise.

To evaluate the proposed method across different calibration scenarios, we compare against three baseline EKF variants. These baselines differ in how the UWB calibration parameters, namely the anchor positions $\mbf{r}_{\ell,j}$ and tag offsets $\mbf{r}_{b,i}$ in the UWB measurement model~\eqref{eq:uwb-model}, are obtained. Our baselines are designed to reflect common practical use cases, ranging from nominal CAD-based calibration to imperfect or data-driven calibration. The three baselines are:
\begin{itemize}
 \item \textbf{CAD-EKF}: Anchor positions $\mbf{r}_{\ell,j}$ and tag offsets $\mbf{r}_{b,i}$ are taken directly from CAD measurements.
 \item \textbf{MisCAD-EKF}: Tag offsets are taken from CAD measurements and perturbed by a $1^\circ$ rotation to simulate mild calibration errors.
 \item \textbf{DataCal-EKF}: Anchor positions and tag offsets are initialized using CAD measurements, then subsequently refined by solving a nonlinear least squares calibration problem with Levenberg-Marquardt optimization on training data.
\end{itemize}
For fairness, the calibration parameters in DataCal-EKF are optimized using the same training data as KILO-EKF.

\begin{table}[t]
    \centering
    \footnotesize
    \caption{Dataset Splits for Cross-Validation}
    \label{tab:folds}
\begin{tabular}{|c|c|c|c|}
\hline
\textbf{Fold} & \textbf{Time [s]} & \textbf{Measurements} & \textbf{Name in MILUV~\cite{miluv}} \\
\hline
1 & 146.1 & 14342 & \emph{default_1_circular2D_0} \\
2 & 135.2 & 13568 & \emph{default_1_circular3D_0} \\
3 & 205.2 & 20700 & \emph{default_1_random3_0} \\
4 & 195.1 & 19512 & \emph{default_1_remoteControlLowPace_0} \\
5 & 227.0 & 22534 & \emph{default_1_remoteControlLowPace_0_v2} \\
\hline
\end{tabular}
\end{table}

\subsection{KILO-EKF Implementation}
\label{subsec:kilo-ekf-implementation}
The proposed KILO-EKF shares the same process model, IMU preprocessing, and state propagation as the baseline EKFs described in Section~\ref{sec:baseline_ekf}. This ensures a controlled comparison, as all methods differ only in their treatment of the measurement model. The key distinction of the KILO-EKF lies in replacing the analytical measurement models with lifted linear-Gaussian representations learned from data.

Inspired by the structure of the UWB range model~\eqref{eq:uwb-model}, we restrict the measurement dependence to the pose variables by defining a reduced state as in~\eqref{eq:reduced-state}. Based on this reduced state, we construct a set of handcrafted features,
\begin{gather}
\mbf{h}(\mbf{s}_k^\prime)
=
\bbm
1 &
\vec(\mbf{C}_k)^\top &
\mbf{C}_k^\top \mbf{t}_k &
\mbf{t}_k^\top \mbf{t}_k
\ebm^\top,
\end{gather}
which encode simple geometric relationships commonly used in range-based localization. These handcrafted features are augmented with SERFFs to form the full lifted state via~\eqref{eq:reduced-state-lifted}.

In addition to lifting the state, we also lift measurements as defined in~\eqref{eq:lifted-quantities}. Specifically, for UWB ranges, we use squared ranges as the lifted measurements: $y_{i,j,k} = p_\gamma(\gamma_{i,j,k}) = \gamma_{i,j,k}^2$. This avoids the square-root nonlinearity in~\eqref{eq:uwb-model}, reducing the complexity of the learned model and the number of features required to capture the measurement relationship.

Because each UWB range measurement depends on a specific anchor position and tag offset, we learn separate lifted measurement models for each anchor-tag pair. This results in $6 \times 2 = 12$ UWB measurement models, along with an additional model for the laser height measurement.

\begin{figure}[t]
\centering
\vspace{5pt}
\includegraphics[width=0.95\columnwidth]{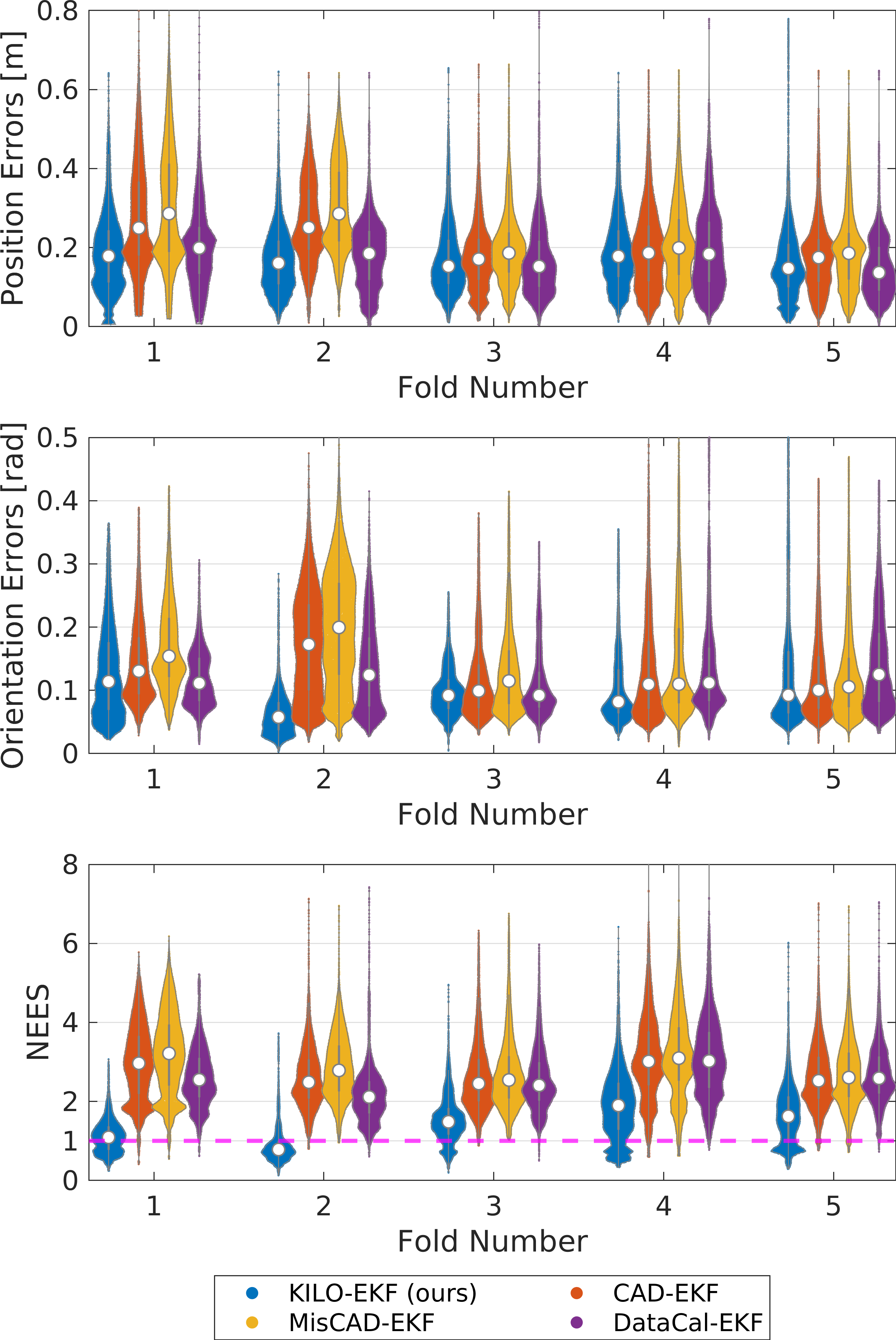}
\caption{Violin plots of per-timestep position errors, orientation errors, and NEES across the five cross-validation folds. Each point corresponds to a single time step. KILO-EKF consistently achieves comparable or lower errors than the baseline methods across all folds, and exhibits NEES values closer to 1, indicating more consistent covariance estimates.}
\label{fig:per_timestep_violins}
\end{figure}
\subsection{Evaluation Method}
We evaluate the proposed KILO-EKF against the three baseline EKF variants. Performance is assessed in terms of position and orientation accuracy and consistency. For quantitative comparison, we report trajectory-level root-mean-square error (RMSE) as well as the normalized estimation error squared (NEES). Lower RMSE indicates higher accuracy, while NEES values close to 1 reflect statistically consistent uncertainty estimates.

Evaluation is performed using a five-fold cross-validation protocol, where one sequence is held out for testing and the remaining four are used for training. The evaluation folds are listed in \tabref{tab:folds}, and visualizations of the trajectories are shown in \figref{fig:koopekf_trajs}. To increase the diversity of training data, we additionally include seven sequences exclusively for training\footnote{Training-only datasets from MILUV~\cite{miluv}: 
\emph{cir_1_circular2D_0},
\emph{cir_3_random3_0},
\emph{default_3_random_0},
\emph{default_3_random_0b},
\emph{default_3_movingTriangle_0b},
\emph{default_3_random2_0},
\emph{default_3_zigzag_0}.}. 
These sequences are excluded from testing, as they are particularly challenging and lead to very large estimation errors for both the proposed method and the baseline EKFs.

\begin{figure}[t]
\centering
\vspace{5pt}
\includegraphics[width=0.95\columnwidth]{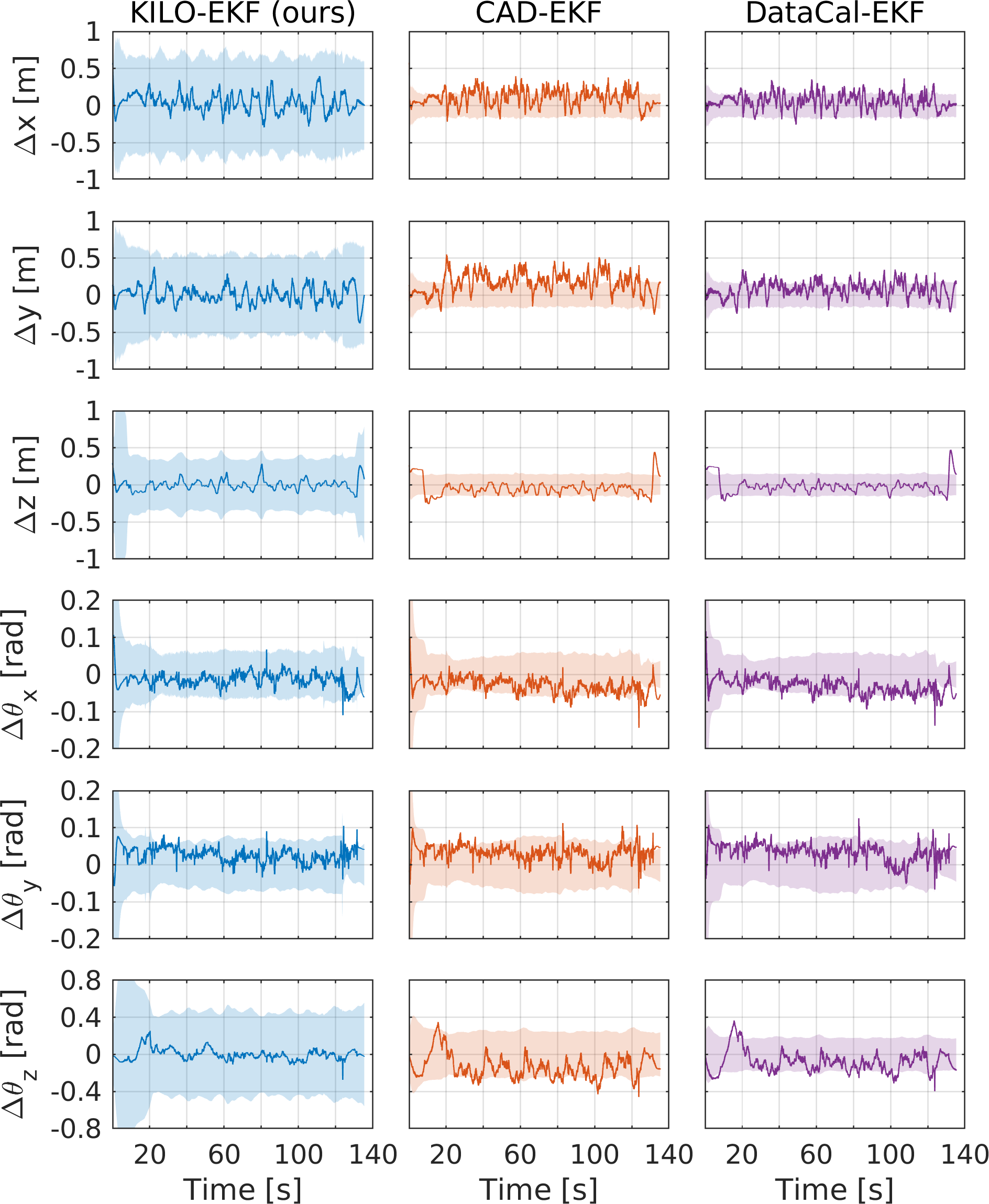}
\caption{Per-axis position and orientation error time series for Fold 2. The shaded region indicates the $3\sigma$ covariance envelope. KILO-EKF exhibits lower errors and more consistent covariance estimates, while the baseline methods show larger errors that frequently exceed the predicted covariance bounds.}
\label{fig:error_plots}
\end{figure}

\begin{figure}[t]
\centering
\vspace{2pt}
\includegraphics[width=0.95\columnwidth]{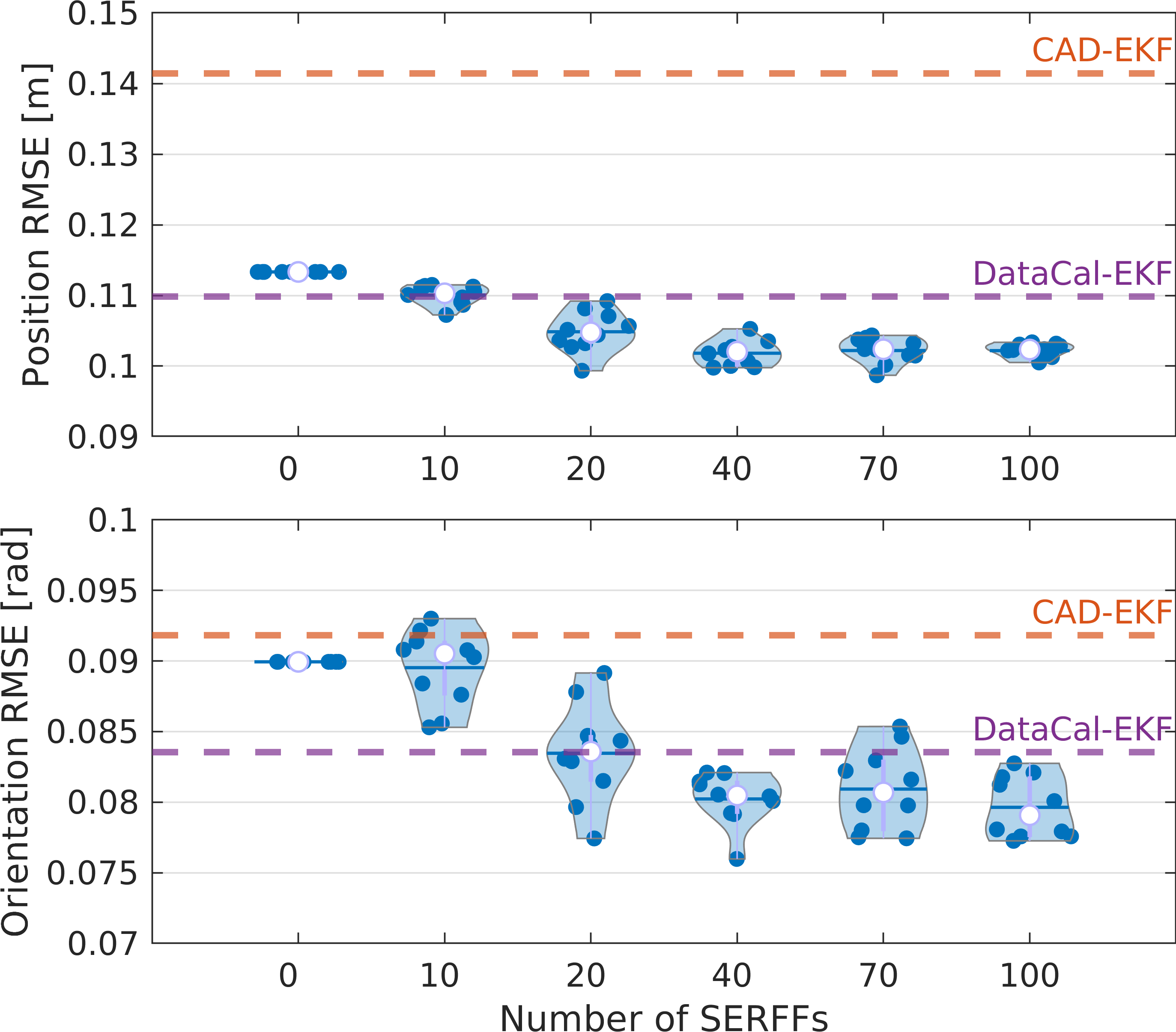}
\caption{Ablation study on SERFFs: KILO-EKF's RMSE versus the number of SERFFs, using the full training dataset. Each dot shows the average RMSE across all folds. Increasing the number of SERFFs consistently reduces error, with KILO-EKF outperforming both baseline methods at 100 SERFFs.}
\label{fig:rmse_vs_rff}
\end{figure}

\begin{figure}[h]
\centering
\includegraphics[width=0.95\columnwidth]{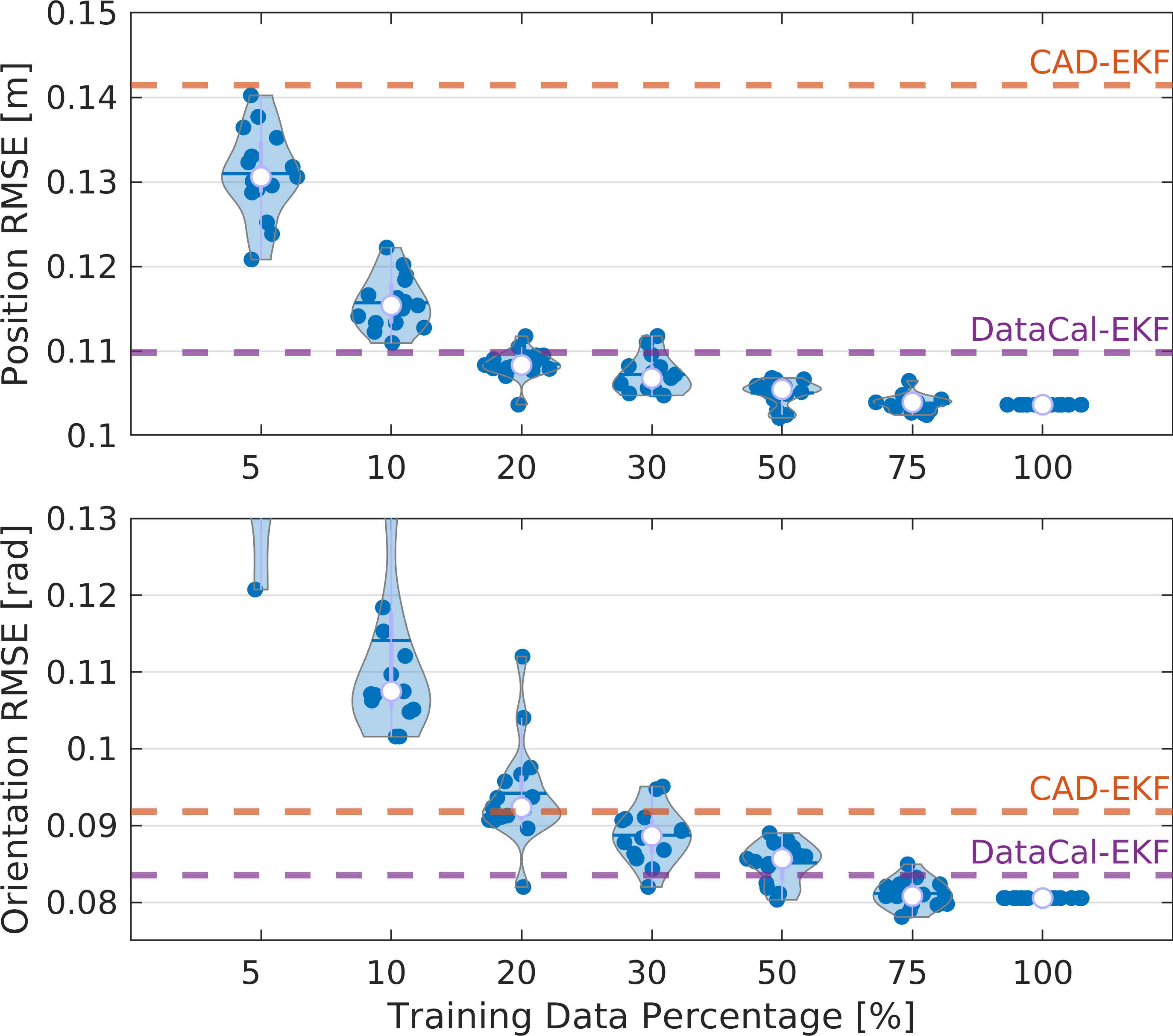}
\caption{Ablation study on training data: KILO-EKF's RMSE versus the percentage of training samples, with the number of SERFFs fixed at 100. Here, DataCal-EKF is trained with the full dataset. Each dot shows the average RMSE across all folds. Increasing the amount of training data consistently reduces error, with KILO-EKF outperforming both baseline methods when using the full dataset.}
\label{fig:rmse_vs_train}
\end{figure}

\begin{figure}[t]
\centering
\vspace{2pt}
\includegraphics[width=0.95\columnwidth]{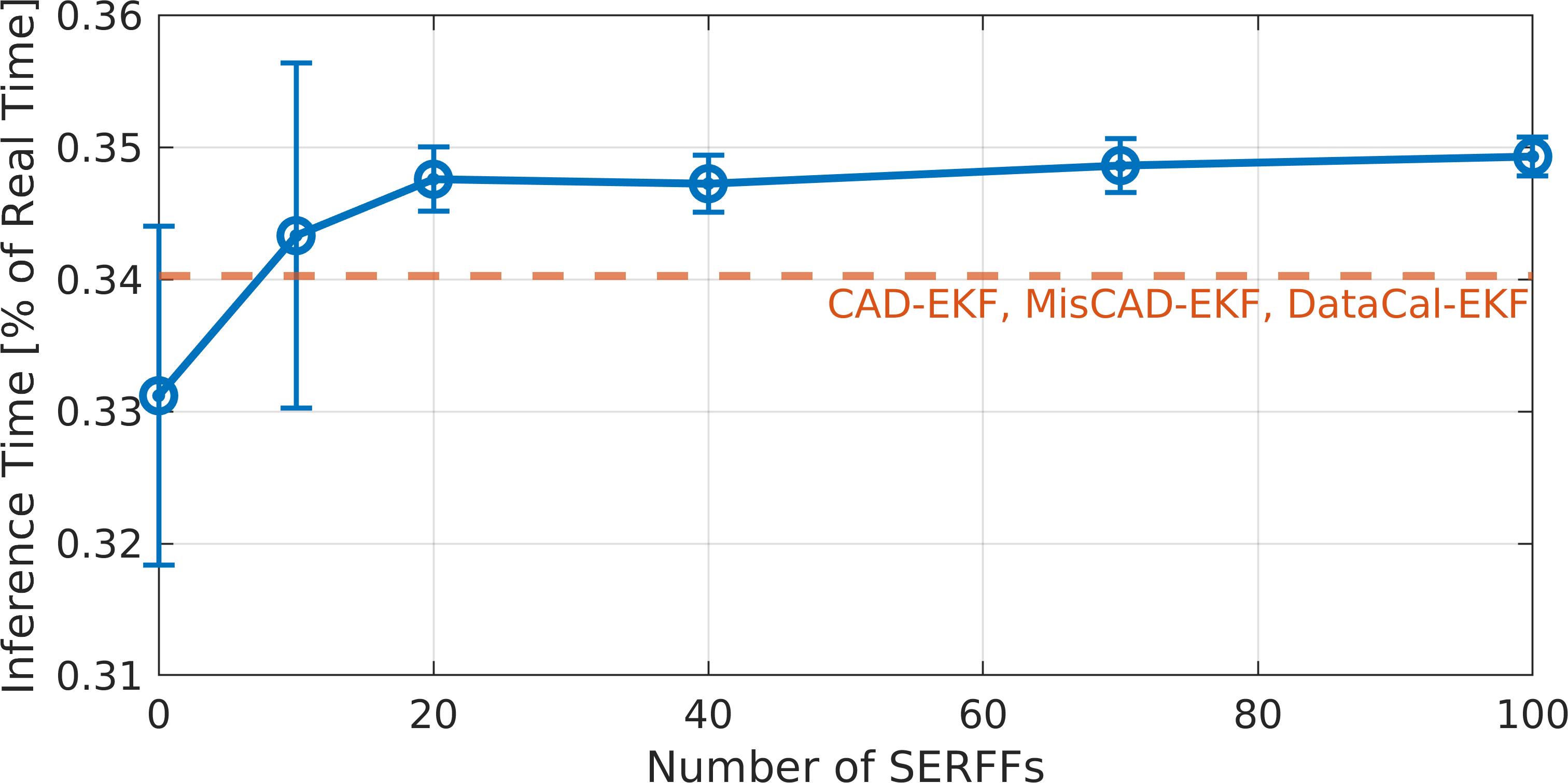}
\caption{KILO-EKF inference time versus the number of SERFFs. Despite operating in a higher-dimensional space, KILO-EKF maintains real-time performance. Its inference time remains comparable to the three baselines, which all have similar inference times, even at 100 SERFFs.}
\label{fig:inference_time_vs_rff}
\end{figure}

\begin{figure}[h]
\centering
\includegraphics[width=0.95\columnwidth]{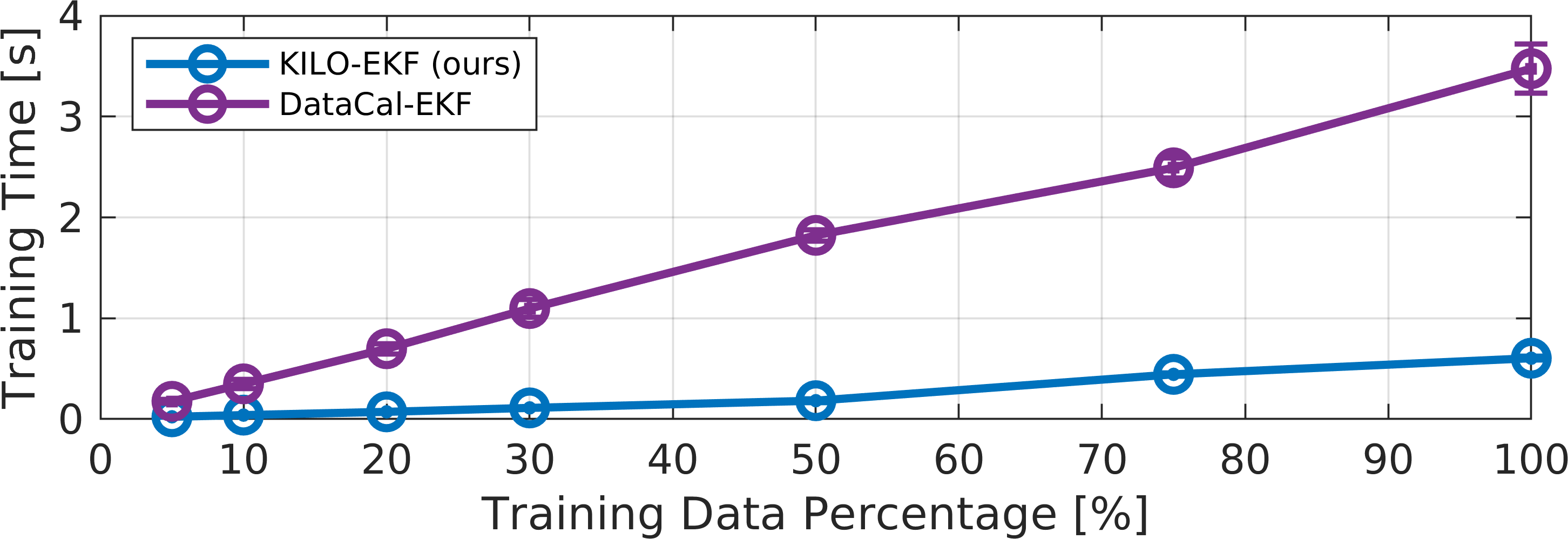}
\caption{Training time versus the percentage of training data. KILO-EKF, which trains in a single pass, scales more favourably than DataCal-EKF, which relies on nonlinear least-squares calibration. As a result, KILO-EKF fits the full dataset (about 200,000 measurements from roughly 30 minutes of data) in under one second.}
\label{fig:training_time_vs_train}
\end{figure}

\subsection{Results}
We compare the proposed KILO-EKF against the three baseline EKF variants using the evaluation metrics described above across the five folds. The distribution of per-timestep errors and NEES values is summarized in \figref{fig:per_timestep_violins}, and the error plots for Fold~2 are shown in \figref{fig:error_plots}. Overall, CAD-EKF and MisCAD-EKF exhibit larger errors and poorer consistency than the two data-calibrated approaches, highlighting the sensitivity of analytical measurement models to calibration accuracy. In particular, MisCAD-EKF demonstrates that even mild perturbations in the assumed tag or anchor geometry lead to noticeable performance degradation.

Comparing the two data-driven approaches, KILO-EKF achieves performance that is consistently comparable to, and in many cases better than, DataCal-EKF across all metrics. While both methods reduce the impact of miscalibration, DataCal-EKF remains constrained by the parametric structure of the analytical measurement model. In contrast, KILO-EKF learns a functional representation of the measurement model, allowing it to capture additional nonlinearities and systematic effects beyond geometric calibration alone. Depending on the fold and metric, this manifests as lower position error, improved orientation accuracy, or more consistent covariance estimates, while remaining competitive in the remaining dimensions.

To further examine these trends, we conduct two ablation studies. First, \figref{fig:rmse_vs_rff} shows the impact of varying the number of SERFFs. When no random features are used, the model relies solely on handcrafted structure and exhibits limited accuracy. As the number of SERFFs increases, the error consistently decreases, and at 100 features, KILO-EKF outperforms both baseline methods. This verifies that capturing additional nonlinearities beyond what is encoded by the analytical model and handcrafted features alone is essential for this problem. Second, \figref{fig:rmse_vs_train} evaluates the effect of training data size. Increasing the amount of training data steadily improves performance, and with the full 30-minute dataset, KILO-EKF surpasses both baseline methods, highlighting the importance of data diversity for learning accurate models. In both cases, improvements exhibit diminishing returns, suggesting a graceful trade-off between model complexity, data volume, and accuracy.

Finally, we evaluate computational efficiency. The experiments are CPU-based implementations run on an Intel Core i7-9750H Processor. As shown in \figref{fig:inference_time_vs_rff}, KILO-EKF maintains real-time inference performance even with a large number of SERFFs, with runtimes comparable to the baseline EKFs. More notably, \figref{fig:training_time_vs_train} demonstrates that KILO-EKF scales favorably in training, fitting the full dataset in under one second due to its closed-form, least-squares-like formulation. This contrasts with DataCal-EKF, which relies on iterative nonlinear optimization and incurs significantly higher computational cost. Although DataCal-EKF converges reliably in our experiments, such nonlinear calibration procedures can also be susceptible to local minima in more complex settings. In contrast, KILO-EKF requires no initialization points and admits a single-pass solution for training.

\section{Conclusion}
\label{sec:conclusion}
We have presented KILO-EKF, a Koopman-inspired extension of the EKF in which the measurement model is learned from data using a linear representation in a lifted feature space. By retaining the standard process model and modifying only the measurement update, KILO-EKF naturally supports Lie-group state spaces such as $\mathrm{SE}_2(3)$. At the same time, it preserves the structure and efficiency of classical filtering while enabling flexible, data-driven modeling of sensor behavior. We have validated our approach on a real-world quadrotor dataset against multiple EKF baselines. The results demonstrate the effectiveness of learned measurement models and the efficiency of the lifted linear formulation, while maintaining real-time performance.

Several directions for future work merit interest. These include exploring structured feature selection methods such as Sparse Identification of Nonlinear Dynamics (SINDy)~\cite{sindy,koop-inv-with-sindy}, extending the framework to support online adaptation of the measurement model, and applying it to broader recursive estimation problems including filter-based SLAM.
\section*{Acknowledgment}
% Acknowledgments redacted for anonymous review.
This work was supported by the Natural Sciences and Engineering Research Council (NSERC) of Canada. We thank Nicholas Dahdah for assistance on the MILUV dataset.

\bibliographystyle{IEEEtran}
{
\singlespacing
\bibliography{refs}
}
\newpage

\end{document}